# Exploring the impact of spatiotemporal granularity on the demand prediction of dynamic ride-hailing

Kai Liu, Zhiju Chen, Toshiyuki Yamamoto and Liheng Tuo

*Abstract*—Dynamic demand prediction is a key issue in ride-hailing dispatching. Many methods have been developed to improve the demand prediction accuracy of an increase in demand-responsive, ride-hailing transport services. However, the uncertainties in predicting ride-hailing demands due to multiscale spatiotemporal granularity, as well as the resulting statistical errors, are seldom explored. This paper attempts to fill this gap and to examine the spatiotemporal granularity effects on ride-hailing demand prediction accuracy by using empirical data for Chengdu, China. A convolutional, long short-term memory model combined with a hexagonal convolution operation (H-ConvLSTM) is proposed to explore the complex spatial and temporal relations. Experimental analysis results show that the proposed approach outperforms conventional methods in terms of prediction accuracy. A comparison of 36 spatiotemporal granularities with both departure demands and arrival demands shows that the combination of a hexagonal spatial partition with an 800 m side length and a 30 min time interval achieves the best comprehensive prediction accuracy. However, the departure demands and arrival demands reveal different variation trends in the prediction errors for various spatiotemporal granularities.

## I. Introduction

As an information and communication technology innovation application in smart transportation, ride-hailing services have become increasingly popular in the daily travel of residents [1]. Although ride-hailing services have improved people's travel quality and provided more mobility options, the dispatching of ride-hailing based on dynamic demand prediction, which helps to reduce the amount of vacant driving and the waiting time, remains an unsolved and urgent issue [2]. The core of ride-hailing services is to connect drivers and passengers with relatively higher spatiotemporal efficiency [3]. Better temporal and spatial prediction of passenger demand is the key to optimizing the dispatching of ride-hailing, to reducing the operating costs of drivers, and to improving the travel quality of residents.

The problem of supply-demand imbalance with high dispersion travel demands has become increasingly prominent for many demand-responsive travel service providers worldwide [4]. As the world's largest ride-hailing service platform with 550 million registered users in 2018, DiDi Chuxing currently responds to more than 50 million orders per day in more than 400 cities. Deep-learning-based methodologies, as well as the dispatching of ride-hailing, have been suggested to be a potential solution for travel demand prediction with this large number of orders, as they address various uncertainties, such as spatial correlation [5], missing data [6], and travel pattern recognition [7].

To estimate and predict the demands of a homogeneous spatial area, a time series model is a good choice if spatial independence exists. However, the demand-responsive services that are available to meet the personalized travel demand exhibit spatial dependence effects, spatial spillover effects, and temporal dependence effects [8], [9]. Another stubborn problem faced by the aggregation of area-based output is the modifiable areal unit problem (MAUP) [10], in which the sensitivity of analytical results to the scale of units is shown to produce highly unreliable results and thus to be essentially unpredictable [11]. They stated that there is no optimal zoning system for all variables from the prospect of minimizing spatial autocorrelation. Although there have been many attempts to address the effects of MAUP, limited successes have been achieved to date [12]. Previous studies on the spatial size and time interval of grid partitions were usually based on experience (involving administrative division, traffic zones and unified grids, such as triangles, squares and hexagons), and no formal

This paragraph of the first footnote will contain the date on which you submitted your paper for review. This work was supported in part by the National Natural Science Foundation of China under Grant 51378091 and Grant 71871043. The authors would like to acknowledge the GAIA open data from DiDi Chuxing. (Corresponding author: Kai Liu.)

K. Liu and Z. Chen are with the School of Transportation and Logistics, Dalian University of Technology, Dalian 116024, China (e-mail: liukai@dlut.edu.cn; chenzhiju@mail.dlut.edu.cn).

T. Yamamoto is with the Institute of Materials and Systems for Sustainability, Nagoya University, Nagoya 464-8603, Japan (e-mail: yamamoto@civil.nagoya-u.ac.jp).

L. Tuo is with Didi Chuxing, Beijing 100085, China (e-mail: tuoliheng@didiglobal.com).

standards or guidelines exist for choosing an appropriate spatiotemporal granularity [13]. However, the ride-hailing demand aggregated by different spatiotemporal granularities have their own regularity. The regularity of the demand volume at the grid level was especially high at times, which indicates that prediction can achieve high accuracy [14]. To the best of our knowledge, the effects of various spatiotemporal granularities on the accuracy of a ride-hailing demand prediction model have often been disregarded.

In this paper, we analyze the multiscale spatiotemporal difference in travel demand prediction of internet-based ride-hailing and propose a hexagonal deep learning approach to examine the effects of spatiotemporal granularity on the prediction accuracy of short-term demands. We addressed the issues of 1) whether the hexagonal deep learning model has better prediction accuracy than a square partition; 2) whether an optimal spatiotemporal granularity exists with better prediction accuracy; and 3) whether the departure demands (how many orders leave at a given time interval in a hexagonal partition) and arrival demands (how many orders arrive at a given time interval in a hexagonal partition) have the same sensitivity to spatiotemporal granularity. Since the spatial positions between two hexagonal partitions are non-Cartesian coordinate data, we propose a convolutional long short-term memory model combined with a hexagonal convolution operation (H-ConvLSTM) to capture complex spatiotemporal correlations.

The remainder of this paper is organized as follows: Section 2 reviews the conventional methods and recent innovations in taxi demand forecasting. Section 3 describes the hexagonal partitions and hexagonal convolution operation and further explicitly introduces the proposed H-ConvLSTM model. An experimental study is conducted in the following section. The last section concludes the paper.

## II. Literature Review

Scientific issues in the network design of taxi services, as well as customer origin-destination (OD) spatiotemporal distributions, have been investigated for more than three decades [15]. Compared to conventional studies on taxi services, the current taxi market, with both roadside pick-up and e-hailing services, induces many interesting and emergent issues. Wang *et al.* [3] proposed a general framework and helped clarify the operation of ridesourcing systems and the complicated interactions among various stakeholders. They stated that four groups of problems must be addressed: 1) demand and pricing, 2) supply and incentives, 3) platform operations, and 4) competition, impacts and regulations.

To improve the prediction accuracy of ridesharing choices, Chen *et al.* [16] examine the service model of an emerging urban mobility company, DiDi Chuxing, and propose a novel ensemble learning approach for understanding on-demand travel patterns. How to effectively mine the potential spatiotemporal characteristics of data and to provide more accurate demand prediction have become increasingly concerning issues in the field of intelligence on demand traffic management. The widely utilized methodology to address spatiotemporal demand estimation includes 1) statistics and econometrics and 2) machine learning.

In addition to the commonly employed statistical methods, such as the moving average, the time series method represented by the autoregressive integrated moving average (ARIMA) model is widely employed in traffic demand prediction [17]. Jiang *et al.* [18] developed a hybrid demand prediction approach for high-speed rail that combines ensemble empirical mode decomposition and gray support vector machine models. Moreira-Matias *et al.* [19] combined three time series forecasting techniques to generate a prediction, the results of which were always better than those obtained by a single model. Ma *et al.* [20] proposed an interactive, multiple-model-based, pattern hybrid approach to predict short-term passenger demands, which maximizes the effective information content by assembling knowledge from pattern models. Davis *et al.* [21] proposed a multilevel clustering technique to improve the accuracy of linear time series model fitting.

Some recent studies found that some unavoidable issues arise when using statistics and econometric methods to predict OD demands, such as the missing spatiotemporal data problem [6], inaccurate analysis of travel demand prediction [22], and especially the ride-hailing matching failure problem due to spatiotemporal rationing discrimination [23]. Travel demand usually has a strong spatiotemporal correlation. From a temporal perspective, affected by factors such as rush hours, weekends and holidays, the travel demand of passengers has a strong periodicity. From a spatial aspect, Yang *et al.* [24] revealed that passenger demand for a particular region is determined jointly by variables for this region and the entire network. Most residents usually have relatively fixed origin points for their travel demands, which may be served by vacant cars from neighboring zones. Therefore, the travel demands of a zone are usually more related to its spatially nearby regions.

With the success of deep learning technology in computer vision and natural language processing [25], [26], the corresponding methods have gradually been applied to traffic prediction [27]. The convolutional neural network (CNN), as a successful deep learning methods, addresses the problem of obtaining spatial relations by treating urban traffic as an image. Ma *et al.* [28] proposed a CNN-based method to improve the accuracy of predicting large-scale, network-wide traffic speeds. Deep learning has also shown advantages in traffic flow prediction [29], OD demand prediction [30], individual mobility prediction, and traffic accident detection.

As one of the best sequence learning methods, long short-term memory (LSTM) has good performance in traffic prediction. Yu *et al.* [31] applied deep LSTM to forecast peak-hour traffic and managed to identify unique characteristics of the traffic data. Xu *et al.* [32] combined LSTM and mixture density networks to predict taxi demand in different areas of a city. Although the LSTM model effectively solves the gradient disappearance problem of long time series by setting the gate unit, they do not model the complex spatiotemporal relationships.

Recent studies proposed the CNN+LSTM model, which combines complex, nonlinear, spatial and temporal correlations to improve the accuracy of the short-term prediction of travel demand and travel time [33], [34]. This model typically uses the CNN model to extract the spatial characteristics of data and then applies the output results as the input of the LSTM model to extract the temporal characteristics, which causes a certain degree of loss in the spatial topological relations of the data. As a solution, Shi *et al.* [35] developed a convolutional LSTM (ConvLSTM) model to address the complex temporal and spatial correlation issues in precipitation nowcasting, which provides convolutional structures in transitions and outperforms the fully connected LSTM. ConvLSTM achieves good performance in the field of video detection [36]. In the domain of transportation, Yang *et al.* [37] utilized travel speed as the input of the ConvLSTM to predict the travel state of a network, and Ke *et al.* [38] and Wang *et al.* [39] applied this model to travel demand prediction.

A limited number of previous studies have explored the shape and scale of spatial partitions, as well as the duration of time partitions, and their effects on the accuracy of transport demand forecasting. Most existing studies divide a city into square grids and then consider the historical spatiotemporal characteristics in each grid as the input for demand forecasting. Otherwise, they choose the traffic area zones based on planners' experience as the spatial statistical unit, which are usually too large and rough in rural-urban continuum areas. Compared with a square, the shape of a hexagon is more similar to a circle, and the points on the boundary are closer to the center, which may be better for grouping similar travel demands into the same partition. In addition, each side of a hexagon is adjacent to a hexagon, which is symmetric equivalent [40]. Because of these advantages, hexagonal partitioning has been widely employed in regional science research. Shoman *et al.* [41] conducted a comparative analysis of the gridding systems of triangles, squares and hexagons for urban land use analysis and discovered that hexagonal grids are the best choice for minimizing the area errors of urban fabric. Csiszár *et al.* [42] applied hexagonal partitioning to the configuration evaluation of urban regional charging stations and further optimized the distribution of charging stations. To the best of our knowledge, Ke *et al.* [43] is the first researcher to propose a CNN model based on hexagonal partitions to address the temporal and spatial correlation issues for supply-demand prediction of ride-sourcing services. However, the local hexagonal partition map is mapped to different matrices through three coordinate systems before the convolution calculation, where the spatial topological relations between each pair of hexagonal partitions suffer from different degrees of loss.

In the field of computer science, efforts to reduce the loss of hexagonal topology have been taken seriously. Steppa *et al.* [44] developed a deep learning framework named HexagDLy with convolution and pooling operations on hexagonal partitions. However, it is difficult to grasp the complex temporal correlation in time series data. Based on the statistical analysis of the taxi OD trajectory dataset, Zhang *et al.* [45] proposed that the choice of the spatial scale has a significant impact on the collected spatial data and the corresponding analysis results. Cheng *et al.* [46] developed a multiscale spatiotemporal anomaly detection model of human activities represented by time series to improve the detection effect and interpretability. In terms of travel demand prediction, Chu *et al.* [47] formed a hierarchical structure such that the output of a larger-scale ConvLSTM is fed to the input of the next level for higher prediction performance. However, very limited research has addressed the scale differences of spatiotemporal partitions and influence mechanisms. To the best of our knowledge, there is only one exception academic paper by Zhang *et al.* [44], which examines the scale effect (granularity effect) of spatial distribution data from the viewpoint of interaction patterns. The scale effect was noticed first by Openshaw and Taylor [48], and researchers have proven that the choice of spatial units may have a significant influence on the results of geographical analysis [49]. The best scale/unit remains unsolved due to the complexity in the spatial interrelation among various spatial activity participants [45], [50], [51]. The current work attempts to understand the scale effect on dynamic ride-hailing demands from the aspect of geoartificial intelligence.

This paper aims to address the research gap of the effect of various spatiotemporal granularities on data-driven, short-term, transport demand forecasting by using an improved topological loss-free H-ConvLSTM model, which combines spatiotemporal prediction with the advantages of hexagons in region partitioning for the travel demand prediction of ride-hailing.

## III. METHODOLOGY

### A. Preliminary

The spatial and temporal distribution characteristics of historical orders are employed to predict the future demand for ride-hailing in a specific partition of a city. Due to the previously described advantages, an urban space is divided into hexagonal partitions as $L = \{l_1, l_2, ..., l_i, ... l_n\}$, and a whole day is discretized into time intervals as $T = \{T_1, T_2, ..., T_t, ... T_m\}$.

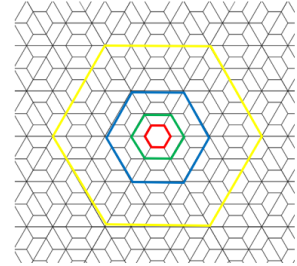

Fig. 1. Hexagonal partitions with various scales

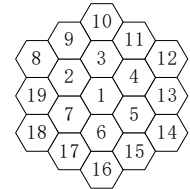

Fig. 2. Hexagonal partition index.

Thus, the demand could be defined as the number of ride-hailing orders at one zone per time interval, i.e., $y_i^t$, which is the demand of location $l_i$ in time interval $T_t$. Because the travel demand of a zone is usually more related to its spatially nearby regions, the surrounding two "rings" of neighbors are simultaneously considered. A hexagonal partition with two layers of its neighbors can be numbered, as shown in Fig. 2. The local hexagon demands centralized at hexagon $l_i$ in time interval $T_t$ are represented as vector $V_i^t = (y_{i_1}^t, y_{i_2}^t, \ldots, y_{i_{19}}^t)$, where $y_{i_1}^t = y_i^t$.

In the temporal aspect, the travel demand of a hexagonal partition is most affected by its past adjacent time intervals. Therefore, the ride-hailing demand prediction model of the target hexagonal partition is established as follows:

$$\hat{y}_i^{t+1} = \mathcal{F}(V_i^{t-h+1}, V_i^{t-h+2}, \ldots, V_i^t) \quad (1)$$

for $i \in L$, where $V_i^{t-h+1}, V_i^{t-h+2}, \ldots, V_i^t$ denotes the historical local hexagon demands for partitions $l_i$ and time intervals from $t - h + 1$ to $t$. $\mathcal{F}(\cdot)$ is the H-ConvLSTM function, which is presented in the following section.

*B. Hexagonal convolution operation*

A hexagonal convolution kernel is adopted to carry out the convolution operation, as shown in Fig. 3. The input and output of a convolution operation are two-layer, local hexagon maps with the same dimensions. The shape distribution of the hexagonal convolution kernel is similar to that of the local hexagonal map. A hexagonal convolution kernel with a kernel size of 1 includes one hexagon and its six adjacent neighbors. Since the forward and feedback propagation of deep learning are based on matrix transformation, both the local hexagonal map and hexagonal convolution kernel need to be matched into the equivalent matrix or tensor. First, the local hexagon map is padded with virtual hexagons of zero demand values to make the data amount of each column equal. Second, the input of the convolution operation can be transformed into a 5×5 2D matrix $X_i^t$, which is established as follows:

$$X_i^t = \begin{bmatrix} 0 & y_{i_9}^t & y_{i_{10}}^t & y_{i_{11}}^t & 0 \\ y_{i_8}^t & y_{i_2}^t & y_{i_3}^t & y_{i_4}^t & y_{i_{12}}^t \\ y_{i_{19}}^t & y_{i_7}^t & y_{i_1}^t & y_{i_5}^t & y_{i_{13}}^t \\ y_{i_{18}}^t & y_{i_{17}}^t & y_{i_6}^t & y_{i_{15}}^t & y_{i_{14}}^t \\ 0 & 0 & y_{i_{16}}^t & 0 & 0 \end{bmatrix} \quad (2)$$

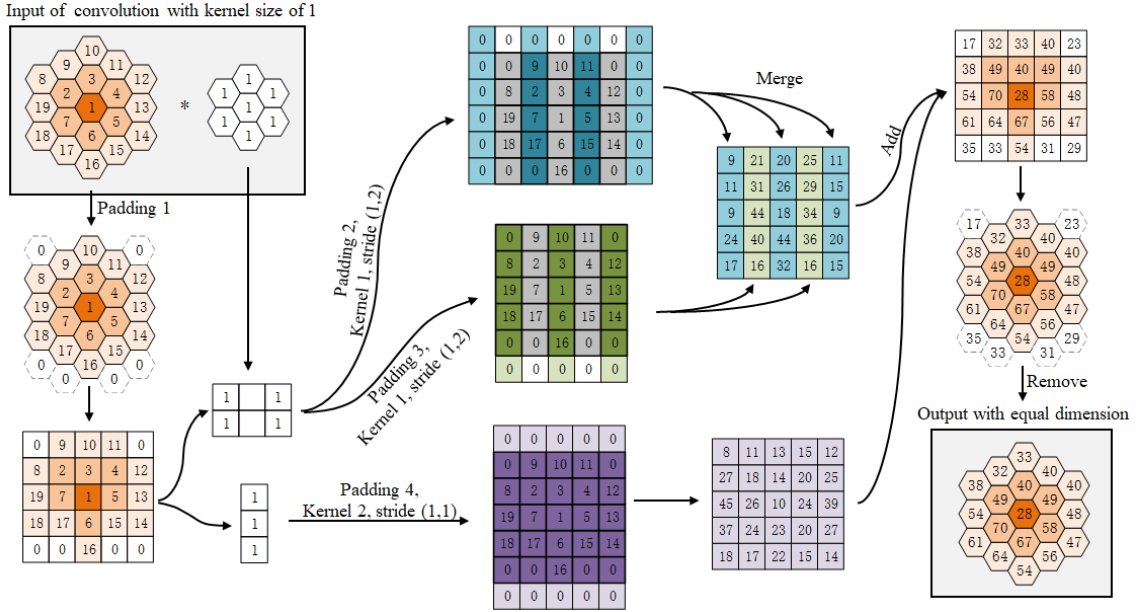

Fig. 3. Realization process of the hexagonal convolution operation with a kernel size of 1.

The hexagonal convolution kernel is decomposed into several rectangular subkernels and combined with the transformed rectangle input tensor for multiple convolutions to realize a hexagonal convolution operation. The specific calculation process follows previous research [44], and a 5×5 2D combinatorial matrix is obtained. Last, remove the data in the redundant position, and output the local hexagon map with equal dimensions as the input.

*C. Architecture of H-ConvLSTM*

Fig. 4 shows the architecture of the H-ConvLSTM model. Assume that the previous $h$ time intervals of historical travel demand of local hexagon maps centralized at hexagon $l_i$ are $V_i^{t-h+1}, \ldots, V_i^t$. As an improved form of the LSTM model, H-ConvLSTM has hexagonal convolutional structures in both the input process and state transfer process and achieves good

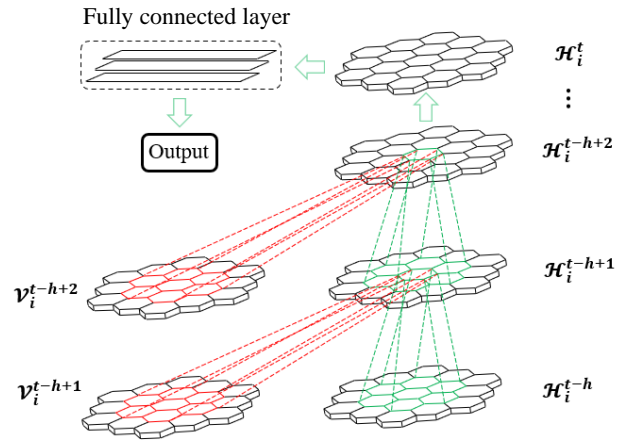

Fig. 4. Architecture of H-ConvLSTM.

performance on general spatiotemporal sequence forecasting problems.

Similar to LSTM, the key to H-ConvLSTM is the cell state, which ensures the memory and circulation of information. H-ConvLSTM has the ability to remove or increase information to the cell state by carefully designing structures that are referred to as gates that consist of forget gates, input gates and output gates. To reduce the loss of spatial topological information in the process of model calculation, the inputs $\mathcal{V}_i^{t-h+1}, \ldots, \mathcal{V}_i^t$, cell states $\mathcal{C}_i^{t-h+1}, \ldots, \mathcal{C}_i^t$, hidden states, $\mathcal{H}_i^{t-h+1}, \ldots, \mathcal{H}_i^t$ and other gates of H-ConvLSTM are 3D tensors whose last two dimensions are rows and columns of spatial information. First, the 2D transformation matrix therefore needs to be transformed to a 3D tensor: $V_i^t \to \mathcal{V}_i^t$. The second step is to determine what information we discard from the cell state $\mathcal{C}_i^t$. This decision is made through the forget gate layer $f_i^t$. The third step is to determine what new information is stored in the cell state. This step consists of an input gate layer $i_i^t$ to determine what information to input and a tanh layer to create a new candidate cell state $\tilde{\mathcal{C}}_i^t$. The old cell state $\mathcal{C}_i^{t-1}$ is then updated to $\mathcal{C}_i^t$. In the last step, we need to determine the output $\mathcal{H}_i^t$, which is based on our new cell state $\mathcal{C}_i^t$ and an output gate layer $\mathcal{O}_i^t$. The output gate layer $\mathcal{O}_i^t$ determines which parts of the cell state $\mathcal{C}_i^t$ will be exported. H-ConvLSTM determines the future state of a certain cell in the grid by the inputs and past states of its local neighbors, which can easily be achieved by using a convolution operator in the input process and state transfer process. The specific functional relationship of H-ConvLSTM is expressed as follows:

$$\begin{aligned}
f_i^t &= \sigma(W_f * [\mathcal{H}_i^{t-1}, \mathcal{V}_i^t] + b_f) \\
i_i^t &= \sigma(W_i * [\mathcal{H}_i^{t-1}, \mathcal{V}_i^t] + b_i) \\
\tilde{\mathcal{C}}_i^t &= tanh(W_c * [\mathcal{H}_i^{t-1}, \mathcal{V}_i^t] + b_c) \\
\mathcal{C}_i^t &= f_i^t \circ \mathcal{C}_i^{t-1} + i_i^t \circ \tilde{\mathcal{C}}_i^t \\
\mathcal{O}_i^t &= \sigma(W_o * [\mathcal{H}_i^{t-1}, \mathcal{V}_i^t] + b_o) \\
\mathcal{H}_i^t &= \mathcal{O}_i^t \circ tanh(\mathcal{C}_i^t)
\end{aligned} \quad (3)$$

where $*$ denotes the hexagonal convolution operator, $\circ$ denotes the Hadamard operator, and $\sigma$ and $tanh$ denote the activation function of the sigmoid function and hyperbolic tangent function, respectively. $W_f, W_i, W_c, W_o, b_f, b_i, b_c, b_i$ are trainable parameters. Given the input $\mathcal{X}_i^t$ and $\mathcal{H}_i^{t-1}$, a 3D tensor $\mathcal{H}_i^t$ that has an integration of temporal and spatial features for location $l_i$ and time interval $T_t$ could be provided.

$\mathcal{H}_i^t$ is input into a fully connected network to obtain the prediction demand $\hat{y}_i^{t+1}$ for location $l_i$ and time interval $T_{t+1}$ as follows:

$$\hat{y}_i^{t+1} = \sigma(W_{fu} \mathcal{H}_i^t + b_{fu}) \quad (4)$$

where $W_{fu}$ and $b_{fu}$ are parameters of the fully connected layer and $\sigma$ is the sigmoid function.

*D. Loss function*

The root mean square error is sensitive to outliers and is usually suitable for the prediction of larger values, while the mean absolute percentage error mainly considers the ratio between the error and the true value and performs well on small-value prediction problems. To combine the advantages of the two error indicators, the loss function is defined as follows:

$$L = \sqrt{\frac{1}{n}\sum_{i=1}^{n}(\hat{y}_i^{t+1} - y_i^{t+1})^2} + \lambda * \frac{1}{n}\sum_{i=1}^{n}\left|\frac{\hat{y}_i^{t+1} - y_i^{t+1}}{y_i^{t+1}}\right| \quad (5)$$

where $\lambda$ is a hyperparameter.

IV. EXPERIMENT

*A. Dataset description*

The dataset is extracted from the DiDi GAIA Open Dataset, which provides the OD information of matched orders and completed trips from Nov. 1st to Nov. 31st, 2016, in Chengdu, China. The total travel demands during this period are approximately 230 thousand orders per day. Fig. 5(a) presents the distribution of the demand for online ride-hailing services throughout the entire month. Friday is the peak time when people choose to leave the city and return home, and leisure shopping travel increases on Saturday; orders placed on these days of the week are 10.8% higher on average compared with other days. The order data of November 15 (Tuesday) and November 19 (Saturday) are selected as the representations of working days and weekends, respectively, in this paper.

The statistics of the travel distribution on both days are shown in Fig. 5(b). The travel demand on both days began to increase sharply at 7 am and gradually decreased after 6 pm. In addition, the travel demand on weekends is generally higher, except during rush hour at approximately 9 am. In this case, additional effort needs to be made by DiDi to fulfil the dynamically changing demand. By using the appropriate prediction model, the online car-hailing supply can be regulated and adjusted in a timely manner according to the spatiotemporal characteristics of the demand.

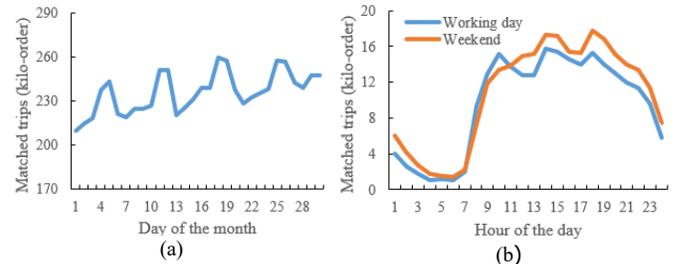

Fig. 5. Relationship between traffic orders and time. (a) Distribution of orders throughout the month. (b) Distribution of orders throughout the day.

A hexagonal grid with a side length of 800 m is adopted to divide the Chengdu urban area, and the time interval is set to 30 min. In addition, the previous 8 time intervals are used to predict the travel demand in the next time interval. Our data include 21×21 hexagonal partitions with longitudes between 115.67°E and 116.01°E and latitudes between 35.73°N and 36.03°N, which cover the majority of urban built-up communities with an area of 733.28 km². The data of the previous 21 days are utilized as the training set, and the data of the remaining 9 days are applied as the test set.

*B. Model configuration*

The experiments are implemented on the DiDi cloud platform with an NVIDIA P4 GPU. The proposed model was conducted via Python 3.6 with PyTorch, TensorFlow and Keras. The size of the local hexagonal cluster is set as a hexagonal partition with two rings of surrounding neighbors, which contains 19 partitions, and the entire research area covers approximately 31.56 km². The proposed H-ConvLSTM model includes 4 ConvLSTM layers, which have 8, 16, 32, and 32 hidden states. The hexagonal kernel size of each layer is 1. Batch normalization and dropout are applied to train the model. The training epochs are set to 50, with a batch size of 128. Adam is used for optimization. The RMSE and MAPE are employed to evaluate the prediction performance. In the loss function, the hyperparameter $\lambda$ is set to 0.01 to ensure that RMSE and MAPE's influence in the feedback calculation is in the same dimension.

*C. Comparison of models*

Several conventional or state-of-the-art models were selected for comparison with our model. To ensure fairness, the two-layer local hexagonal map $V_i^t$ centered on the specific hexagonal partition $y_i^t$ is selected as the input features of each benchmark model. In terms of the partition shape, the square partition with the same area as the hexagon is selected as a contrast, and the input of the corresponding benchmark models is a 5×5 local adjacency matrix centered on $y_i^t$. The configuration of each benchmark model is presented as follows:

1) Historical Average: indicates that the demand value is periodic and that the weighted average value of the demand in the historical period was employed to predict the demand in the next period.

2) ARIMA: the ARIMA model that is widely utilized for time series prediction. The difference order $d$ is set to 1, with autoregressive coefficient $p$ and moving average coefficient $q$ iterating previous time intervals between 1 and 8.

3) CNN: the previous 8 time intervals are represented by the number of channels of the input image. Before the convolution calculation, the two-layer local hexagon map should be transformed into a 5×9 matrix following previous research [43]. The CNN model with both hexagonal partitions and square partitions includes 4 convolution layers, which have 8, 16, 32, and 32 hidden states. The convolution kernel size of each layer is 3×3. Batch normalization and dropout are employed to train the model.

4) LSTM: local adjacency maps of hexagon and square partitions are flattened into 1-dimensional demand vectors, and demand features of the previous 8 time intervals are selected to predict the next time interval. The hidden state is set to 128.

5) ConvLSTM: The two-layer, local hexagon map is transformed into a 5×9×1 tensor following previous research [43]. The convolution kernel size of each layer is 3×3. Other parameter settings are consistent with the H-ConvLSTM model.

6) H-CNN: the previous 8 time intervals are represented by the number of channels of the input image. A hexagonal convolution operation is applied between each layer. The hexagonal kernel size of each layer is 1. Other parameter settings are consistent with the CNN model.

The performance results with departure demand prediction of each model are shown in Table I. Compared with the square, the hexagon is symmetric equivalent and more resembles the circle, which may be better for grouping similar travel demands into the same partition. The MAPE and RMSE are improved to a certain extent in each model with hexagonal partitioning. With the enhancement of the ability to capture the temporal and spatial characteristics of the travel demand, the prediction performance of ConvLSTM with hexagonal partition improves by 21.3% and 10.2% compared to the CNN and LSTM, respectively, in terms of the MAPE (16.0% and 2.8%, in terms of the RMSE). Due to hexagonal convolutional structures in both the input process and state transfer process, the proposed H-ConvLSTM model eliminates the spatial topological loss of travel demand caused by matrix transformation in the ConvLSTM model and achieves the best comprehensive prediction accuracy. Moreover, the training set is evenly divided into three parts $P_1$, $P_2$ and $P_3$ according to time, which are respectively used to exchange with the data of the previous week in the test set, as shown in Fig. 6. Another three groups of data combinations $G_1$, $G_2$ and $G_3$ are therefore obtained to verify the statistical stability of the prediction performance. Systematic cross-validation shows that the proposed H-ConvLSTM model has smaller standard deviation. In addition, two-sample t-tests between the proposed H-ConvLSTM model and other models are performed for MAPE and RMSE distributions under each data combination. The hypothesis test between H-ConvLSTM and the second-best model ConvLSTM (with hexagon partition) shows the largest P-value 0.0029, in terms of the MAPE (0.0015, in terms of the RMSE), which indicates that the optimal prediction performance of our proposed model is statistically significant.

Fig. 6. Partitioning modes of data set.

TABLE I
COMPARISON OF VARIOUS DEPARTURE DEMAND PREDICTION MODELS

| Model | Partition shape | RMSE/Sd. | MAPE(×10⁻²)/Sd. | Train time (min) | Test time (min) |
|---|---|---|---|---|---|
| HA | Square | 12.38/0.20 | 27.22/0.50 | 0.03 | 0.01 |
| | Hexagon | 12.17/0.22 | 25.36/0.77 | 0.03 | 0.01 |
| ARIMA | Square | 12.56/0.36 | 23.61/0.61 | 0.12 | 0.01 |
| | Hexagon | 11.93/0.13 | 23.29/0.58 | 0.13 | 0.01 |
| CNN | Square | 11.40/0.14 | 22.45/0.76 | 5.16 | 0.02 |
| | Hexagon | 10.75/0.15 | 21.62/0.49 | 6.71 | 0.02 |
| LSTM | Square | 9.46/0.13 | 20.62/0.68 | 10.02 | 0.03 |
| | Hexagon | 9.29/0.11 | 18.95/0.58 | 10.13 | 0.03 |
| ConvLSTM | Square | 9.37/0.12 | 17.18/0.46 | 47.84 | 0.11 |
| | Hexagon | 9.03/0.07 | 17.02/0.55 | 52.57 | 0.13 |
| H-CNN | Hexagon | 9.64/0.16 | 20.13/0.40 | 127.58 | 0.31 |
| H-ConvLSTM | Hexagon | **8.82/0.05** | **16.71/0.36** | 163.43 | 0.42 |

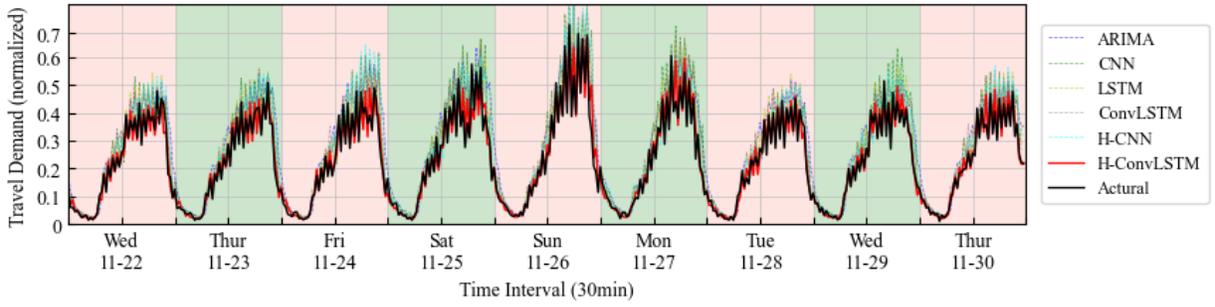

Fig. 7. Demand prediction performance and actual value in a hexagon.

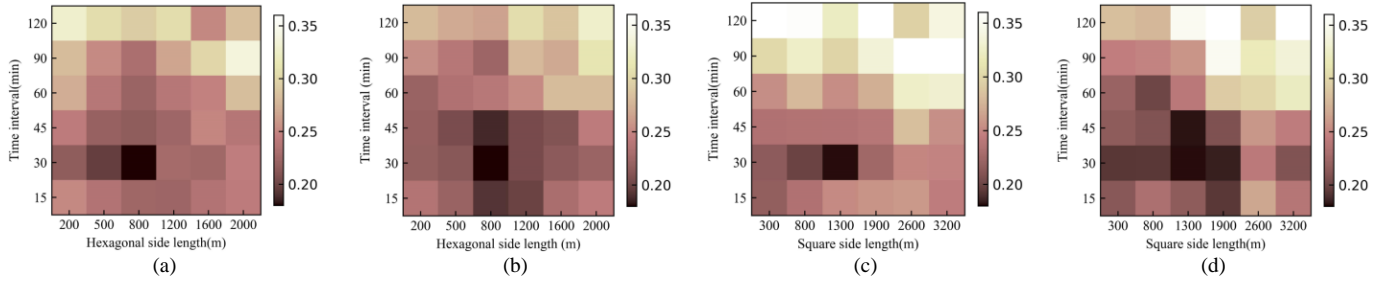

Fig. 8. MAPE of OD prediction at various spatiotemporal granularities. (a) Departure demands for hexagon partition. (b) Arrival demands for hexagon partition. (c) Departure demands for square partition. (d) Arrival demands for square partition.

Since the forward and feedback propagation of deep learning are based on matrix transformation, both the local hexagonal map and hexagonal convolution kernel need to be matched to the equivalent matrix or tensor, which produces a higher training time for our H-ConvLSTM model. However, considering that there is no order of magnitude difference in the training time between our model and the second-best model and that all training is conducted offline, the potential disadvantage of this higher training time to demand prediction is limited. Fig. 7 shows the trend of travel demand prediction and actual results of each model in a selected hexagon over time. The H-ConvLSTM model shows more stable prediction performance in high demand time intervals.

### D. Effects of spatiotemporal granularity

The ideal spatiotemporal granularity is expected to reflect similar OD patterns, which may be influenced by the built environment, as well as travelers' personal socioeconomic characteristics and travel habits. Previous studies on the spatial size division and the time interval selection of the basic research unit usually made judgments by experience, taking, e.g., 500 m or 1 km, as the spatial size and 30 min or 60 min as the time interval, and no unified standard exists. To achieve better prediction performance, the prediction performances for hexagonal partitions and square partitions at various spatiotemporal granularities are compared. The spatial granularities of hexagonal partitions are set to 200 m, 500 m, 800 m, 1200 m, 1600 m, and 2000 m. To ensure consistency in the area of the partitions, the spatial granularities of the corresponding square partitions are set to 300 m, 800 m, 1300 m, 1900 m, 2600 m, and 3200 m. The temporal granularities of the two partition shapes are uniformly set to 15 min, 30 min, 45 min, 60 min, 90 min, and 120 min.

The results of the MAPE shown in Fig. 8 reveal that the intuition judgment of "the smaller a partition (with a higher similar regional build environment) is, the smaller the heterogeneity uncertain effect on the prediction performance will be" is sometimes not the truth. As the granularity of the temporal and spatial dimensions increases, the MAPE of the prediction of both the departure demand and arrival demand for hexagonal partitions and square partitions decreases and then increases. The predicted short-term departure demand and arrival demand have the best comprehensive prediction performance for a time granularity of 30 min and a hexagonal partition with a side length of 800 m (corresponding square partition with a side length of 1300 m) for Chengdu. The scale of 800 m (1300 m) and 30 min for hexagonal partition (square partition) is the best choice for the spatiotemporal granularity, which balances the within-zone similarity and between-zone variance, as travelers with similar travel demand characteristics are not divided into a common partition if the spatiotemporal granularity is too small, which leads to unstable demand of the partition. If the spatiotemporal granularity is too large, more demands with different spatial and temporal demand characteristics are grouped, and the sample size of the data is drastically reduced. For example, the sample size of the training set with a time granularity of 30 min and a hexagonal partition with a side length of 800 m is $21\times21\times21\times48=444528$, while the sample size of the training set with a time granularity of 90 min and a hexagonal partition with a side length of 2000 m is $8\times8\times21\times16=21502$, which has a negative impact on model training and renders the prediction of the model more complicated.

Another interesting phenomenon is that the two types of demands are quite different when discussing the spatial granularity with the second-best prediction accuracy. The departure demand prediction tends to be better with a smaller hexagonal partition side length of 500 m (corresponding square partition side length of 800 m) and the same time interval of 30 min, while the arrival demand prediction performs better at the same spatial partition side length of 800 m (corresponding square partition side length of 1300 m) and a longer time interval of 45 min. To quickly and easily access a ride-hailing car, passengers usually choose a pick-up point that is relatively easy to access to meet the driver. Origin points with similar travel characteristics are therefore relatively fixed and concentrated. However, to reduce the walking distance, passengers often ask their drivers to take them as close as possible to their destination. The distribution of passenger destinations is therefore more discrete and complex and requires a large spatial granularity for analysis.

Generally, the RMSE is an absolute error; thus, the smaller the scale of the spatiotemporal granularity is, the better the accuracy will be. Notably, the RMSE is sensitive to outliers. With an increase in spatial granularity, the demand values of both the hexagonal partitions and square partitions increase significantly. For example, the maximum travel demand of the training set with a time granularity of 30 min and a hexagonal partition with a side length of 800 m is 859, while the maximum travel demand of the training set with a time granularity of 90 min and a hexagonal partition with a side length of 2000 m is 4192. The increased probability of large outliers eventually leads to higher RMSE values. To better reflect the optimal spatial granularity, we normalized this indicator as follows:

$$NRMSE = \frac{RMSE}{y_{max} - y_{min}} \quad (6)$$

where NRMSE is the normalized root mean square error and $y_{max}$ and $y_{min}$ are the maximum value and minimum value, respectively, of travel demand at each spatiotemporal granularity.

The effects of the spatiotemporal granularity on the NRMSE are shown in Fig. 9. When the hexagonal side length is 800 m (corresponding square partition side length of 1300 m) and the time interval is 30 min, the departure demands and arrival demands also have the optimal prediction performance. In the discussion of the spatial granularity with the second-best prediction accuracy, the prediction accuracy gap between the second-best granularity and the other granularities decreases.

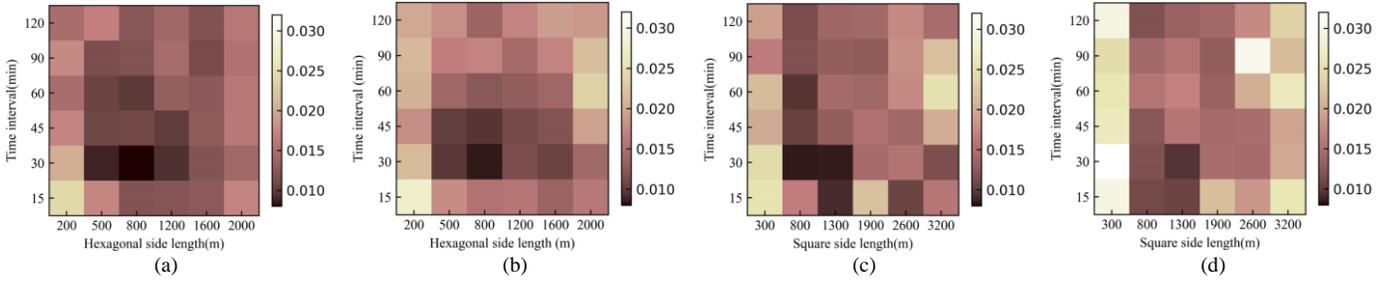

Fig. 9. NRMSE of OD prediction at various spatiotemporal granularities. (a) Departure demands for hexagon partition. (b) Arrival demands for hexagon partition. (c) Departure demands for square partition. (d) Arrival demands for square partition.

## V. CONCLUSIONS

This paper presents an improved deep learning approach to short-term demand prediction of ride-hailing services, which addresses the gap of the disregarded effects of spatiotemporal granularity on the prediction accuracy. The main contributions of this study are presented as follows:

We use hexagon-based zoning to develop an H-ConvLSTM deep learning model to better explore potential spatial and temporal correlations, which is a commonly disregarded issue in statistical- or econometric-based models. The proposed H-ConvLSTM model is a novel combination of the conventional ConvLSTM model and HexagDLy framework, with the advantages of no topology loss through hexagonal convolution operation, which is more complex than both the ConvLSTM and HexagDLy frameworks and has never been proposed in the field of travel demand prediction. Experimental results show that the proposed method has better prediction performance than several competing methods.

We revealed the issue and characteristics of the scale effect on ride-hailing demand prediction. One of the most important findings is that the scale effect cannot be well addressed by widely utilized deep learning methods due to the natural, spatiotemporal, unbalanced characteristics of dynamic travel demands. Therefore, we conducted a sensitivity analysis to analyze the effect of spatiotemporal granularity on demand prediction and addressed the gap of providing guidance on choosing spatial and temporal granularity for accurate demand predictions.

The influence of spatiotemporal granularity on travel OD demand prediction for hexagonal partitions and square partitions is explored considerably by examining 72 situations, including six spatial granularities, six temporal granularities and two types of demand (departure demands and arrival demands). The results verify that spatiotemporal granularity does influence the accuracy of short-term OD prediction, which has similar distribution characteristics for the two partition types. Chengdu achieves better prediction performance when the time interval is 30 min and the side length of the hexagonal partition is 800 m (corresponding square partition with a side length of 1300 m). An interesting finding is that the departure demands and arrival demands have different trends with the second-best spatiotemporal granularity, where departure demands tend to be clustered in a smaller hexagonal partition side length of 500 m (corresponding square partition side length of 800 m) for the same 30 min temporal granularity and arrival demands achieve better accuracy if clustered with a longer temporal granularity of 45 min under the same 800 m spatial granularity (corresponding square partition side length of 1300 m). These findings are useful for supply vehicle dispatching.

The central urban areas with intensive demand need to provide more refined demand prediction in pursuit of supply-demand matching efficiency, while the suburbs with sparse demand need expanded aggregation scale to improve the prediction accuracy. Therefore, designing a travel demand partition division method with multiple spatiotemporal granularities will become an important research direction to get the optimal spatiotemporal granularity and capture the travel demand spatiotemporal patterns of different locations in future work, which will further improve the prediction performance of internet-based ride-hailing. Moreover, we plan to pay more attention to improving the efficiency of model training in future studies.

REFERENCES

[1] L. Zha, Y. Yin, and H. Yang, "Economic analysis of ride-sourcing markets," *Transp. Res. Part C Emerg. Technol.*, vol. 71, pp. 249–266, 2016, doi: 10.1016/j.trc.2016.07.010.

[2] H. Yang, X. Qin, J. Ke, and J. Ye, "Optimizing matching time interval and matching radius in on-demand ride-sourcing markets," *Transp. Res. Part B Methodol.*, vol. 131, pp. 84–105, 2020, doi: 10.1016/j.trb.2019.11.005.

[3] H. Wang and H. Yang, "Ridesourcing systems: A framework and review," *Transp. Res. Part B Methodol.*, vol. 129, pp. 122–155, 2019, doi: 10.1016/j.trb.2019.07.009.

[4] O. Rifki, N. Chiabaut, and C. Solnon, "On the impact of spatio-temporal granularity of traffic conditions on the quality of pickup and delivery optimal tours," *Transp. Res. Part E Logist. Transp. Rev.*, vol. 142, no. September, p. 102085, 2020, doi: 10.1016/j.tre.2020.102085.

[5] Y. Li and B. Shuai, "Origin and destination forecasting on dockless shared bicycle in a hybrid deep-learning algorithms," *Multimed. Tools Appl.*, vol. 79, no. 7–8, pp. 5269–5280, Feb. 2020, doi: 10.1007/s11042-018-6374-x.

[6] X. Chen, Z. He, and L. Sun, "A Bayesian tensor decomposition approach for spatiotemporal traffic data imputation," *Transp. Res. Part C*, vol. 98, no. October 2018, pp. 73–84, 2019, doi: 10.1016/j.trc.2018.11.003.

[7] Y. Dong, S. Wang, L. Li, and Z. Zhang, "An empirical study on travel patterns of internet based ride-sharing," *Transp. Res. Part C Emerg. Technol.*, vol. 86, no. October 2017, pp. 1–22, 2018, doi: 10.1016/j.trc.2017.10.022.

[8] J. Wang, T. Yamamoto, and K. Liu, "Key determinants and heterogeneous frailties in passenger loyalty toward customized buses: An empirical investigation of the subscription termination hazard of users," *Transp. Res. Part C Emerg. Technol.*, vol. 115, no. April, p. 102636, 2020, doi: 10.1016/j.trc.2020.102636.

[9] J. Wang, T. Yamamoto, and K. Liu, "Spatial interdependence and spillover effects in customized bus demand: Empirical evidence using spatial dynamic panel models," 2020.

[10] S. Openshaw, *The modifiable areal unit problem*. Norwich, UK: Geo Books, 1984.

[11] A. S. Fotheringham and D. W. S. Wong, "The modifiable areal unit problem in multivariate statistical analysis," *Environ. Plan. A*, vol. 23, no. 7, pp. 1025–1044, 1991, doi: 10.1068/a231025.

[12] D. W. Wong, "Modifiable Areal Unit Problem," in *International Encyclopedia of Human Geography*, Elsevier, 2009, pp. 169–174.

[13] K. Zhang, Y. Chen, and Y. (Marco) Nie, "Hunting image: Taxi search strategy recognition using Sparse Subspace Clustering," *Transp. Res. Part C Emerg. Technol.*, vol. 109, no. May, pp. 250–266, 2019, doi: 10.1016/j.trc.2019.10.015.

[14] X. Zhang, "Beyond expected regularity of aggregate urban mobility: A case study of ridesourcing service," *J. Transp. Geogr.*, vol. 95, no. May, p. 103150, 2021, doi: 10.1016/j.jtrangeo.2021.103150

[15] H. Yang and S. C. Wong, "A network model of urban taxi services," *Transp. Res. Part B Methodol.*, vol. 32, no. 4, pp. 235–246, 1998, doi: 10.1016/S0191-2615(97)00042-8.

[16] X. (Michael) Chen, M. Zahiri, and S. Zhang, "Understanding ridesplitting behavior of on-demand ride services: An ensemble learning approach," *Transp. Res. Part C Emerg. Technol.*, vol. 76, pp. 51–70, 2017, doi: 10.1016/j.trc.2016.12.018.

[17] B. M. Williams and L. A. Hoel, "Modeling and Forecasting Vehicular Traffic Flow as a Seasonal ARIMA Process: Theoretical Basis and Empirical Results," *J. Transp. Eng.*, vol. 129, no. 6, pp. 664–672, 2003, doi: 10.1061/(ASCE)0733-947X(2003)129:6(664).

[18] X. Jiang, L. Zhang, and M. X. Chen, "Short-term forecasting of high-speed rail demand: A hybrid approach combining ensemble empirical mode decomposition and gray support vector machine with real-world applications in China," *Transp. Res. Part C Emerg. Technol.*, vol. 44, pp. 110–127, 2014, doi: 10.1016/j.trc.2014.03.016.

[19] L. Moreira-Matias, J. Gama, M. Ferreira, J. Mendes-Moreira, and L. Damas, "Predicting Taxi–Passenger Demand Using Streaming Data," *IEEE Trans. Intell. Transp. Syst.*, vol. 14, no. 3, pp. 1393–1402, Sep. 2013, doi: 10.1109/TITS.2013.2262376.

[20] Z. Ma, J. Xing, M. Mesbah, and L. Ferreira, "Predicting short-term bus passenger demand using a pattern hybrid approach," *Transp. Res. Part C Emerg. Technol.*, vol. 39, pp. 148–163, 2014, doi: 10.1016/j.trc.2013.12.008.

[21] N. Davis, G. Raina, and K. Jagannathan, "A multi-level clustering approach for forecasting taxi travel demand," *IEEE Conf. Intell. Transp. Syst. Proceedings, ITSC*, pp. 223–228, 2016, doi: 10.1109/ITSC.2016.7795558.

[22] N. Sanko, T. Morikawa, and Y. Nagamatsu, "Post-project evaluation of travel demand forecasts: Implications from the case of a Japanese railway," *Transp. Policy*, vol. 27, pp. 209–218, 2013, doi: 10.1016/j.tranpol.2013.02.002.

[23] Z. Xu, Y. Yin, and J. Ye, "On the supply curve of ride-hailing systems," *Transp. Res. Part B Methodol.*, vol. 132, no. xxxx, pp. 29–43, Mar. 2020, doi: 10.1016/j.trb.2019.02.011.

[24] H. Yang, C. W. Y. Leung, S. C. Wong, and M. G. H. Bell, "Equilibria of bilateral taxi-customer searching and meeting on networks," *Transp. Res. Part B Methodol.*, vol. 44, no. 8–9, pp. 1067–1083, 2010, doi: 10.1016/j.trb.2009.12.010.

[25] A. Graves, A. R. Mohamed, and G. Hinton, "Speech recognition with deep recurrent neural networks," *ICASSP, IEEE Int. Conf. Acoust. Speech Signal Process. - Proc.*, no. 6, pp. 6645–6649, 2013, doi: 10.1109/ICASSP.2013.6638947.

[26] A. Krizhevsky, I. Sutskever, and G. E. Hinton, "ImageNet classification with deep convolutional neural networks," *Commun. ACM*, vol. 60, no. 6, pp. 84–90, May 2017, doi: 10.1145/3065386.

[27] H. Yi, J. Heejin, and S. Bae, "Deep Neural Networks for traffic flow prediction," *2017 IEEE Int. Conf. Big Data Smart Comput. BigComp 2017*, pp. 328–331, 2017, doi: 10.1109/BIGCOMP.2017.7881687.

[28] X. Ma, Z. Dai, Z. He, J. Ma, Y. Wang, and Y. Wang, "Learning Traffic as Images: A Deep Convolutional Neural Network for Large-Scale Transportation Network Speed Prediction," *Sensors*, vol. 17, no. 4, p. 818, Apr. 2017, doi: 10.3390/s17040818.

[29] N. G. Polson and V. O. Sokolov, "Deep learning for short-term traffic flow prediction," *Transp. Res. Part C Emerg. Technol.*, vol. 79, pp. 1–17, 2017, doi: 10.1016/j.trc.2017.02.024.

[30] L. Liu, Z. Qiu, G. Li, Q. Wang, W. Ouyang, and L. Lin, "Contextualized Spatial–Temporal Network for Taxi Origin-Destination Demand Prediction," *IEEE Trans. Intell. Transp. Syst.*, vol. 20, no. 10, pp. 3875–3887, Oct. 2019, doi: 10.1109/TITS.2019.2915525.

[31] R. Yu, Y. Li, C. Shahabi, U. Demiryurek, and Y. Liu, "Deep Learning: A Generic Approach for Extreme Condition Traffic Forecasting," *Proc. 2017 SIAM Int. Conf. Data Min.*, pp. 777–785, 2017, doi: 10.1137/1.9781611974973.87.

[32] J. Xu, R. Rahmatizadeh, L. Boloni, and D. Turgut, "Real-Time Prediction of Taxi Demand Using Recurrent Neural Networks," *IEEE Trans. Intell. Transp. Syst.*, vol. 19, no. 8, pp. 2572–2581, Aug. 2018, doi: 10.1109/TITS.2017.2755684.

[33] W. Wei, X. Jia, Y. Liu, and X. Yu, "Travel Time Forecasting with Combination of Spatial-Temporal and Time Shifting Correlation in CNN-LSTM Neural Network," in *Cai Y., Ishikawa Y., Xu J. (eds) Web and Big Data. APWeb-WAIM 2018. Lecture Notes in Computer Science*, vol. 10987, Springer, Cham, 2018.

[34] H. Yao *et al.*, "Deep Multi-View Spatial-Temporal Network for Taxi Demand Prediction," *Proc. Thirty-Second AAAI Conf. Artif. Intell.*, Feb. 2018, [Online]. Available: http://arxiv.org/abs/1802.08714.

[35] X. Shi, Z. Chen, H. Wang, D.-Y. Yeung, W. Wong, and W. Woo, "Convolutional LSTM Network: A Machine Learning Approach for Precipitation Nowcasting," *Adv. Neural Inf. Process. Syst.*, vol. 2015-Janua, no. June, pp. 68–80, Jun. 2015, doi: 10.1007/978-3-319-21233-3_6.

[36] M. Majd and R. Safabakhsh, "A motion-aware ConvLSTM network for action recognition," *Appl. Intell.*, vol. 49, no. 7, pp. 2515–2521, 2019, doi: 10.1007/s10489-018-1395-8.

[37] G. Yang, Y. Wang, H. Yu, Y. Ren, and J. Xie, "Short-Term Traffic State Prediction Based on the Spatiotemporal Features of Critical Road Sections," *Sensors*, vol. 18, no. 7, p. 2287, Jul. 2018, doi: 10.3390/s18072287.

[38] J. Ke, H. Zheng, H. Yang, and X. (Michael) Chen, "Short-term forecasting of passenger demand under on-demand ride services: A spatio-temporal deep learning approach," *Transp. Res. Part C Emerg. Technol.*, vol. 85, pp. 591–608, 2017, doi: 10.1016/j.trc.2017.10.016.

[39] D. Wang, Y. Yang, and S. Ning, "DeepSTCL: A Deep Spatio-temporal ConvLSTM for Travel Demand Prediction," in *2018 International Joint Conference on Neural Networks (IJCNN)*, Jul. 2018, vol. 2018-July, pp. 1–8, doi: 10.1109/IJCNN.2018.8489530.

[40] C. P. D. Birch, N. Vuichard, and B. R. Werkman, "Modelling the effects of patch size on vegetation dynamics: Bracken [Pteridium aquilitnum (L.) Kuhn] under grazing," *Ann. Bot.*, vol. 85, no. SUPPL. B, pp. 63–76, 2000, doi: 10.1006/anbo.1999.1081.

[41] W. Shoman, U. Alganci, and H. Demirel, "A comparative analysis of gridding systems for point-based land cover/use analysis," *Geocarto Int.*, vol. 34, no. 8, pp. 867–886, 2019, doi: 10.1080/10106049.2018.1450449.

[42] C. Csiszár, B. Csonka, D. Földes, E. Wirth, and T. Lovas, "Urban public charging station locating method for electric vehicles based on land use approach," *J. Transp. Geogr.*, vol. 74, no. June 2018, pp. 173–180, 2019, doi: 10.1016/j.jtrangeo.2018.11.016.

[43] J. Ke *et al.*, "Hexagon-Based Convolutional Neural Network for Supply-Demand Forecasting of Ride-Sourcing Services," *IEEE Trans. Intell. Transp. Syst.*, vol. 20, no. 11, pp. 4160–4173, Nov. 2019, doi: 10.1109/TITS.2018.2882861.

[44] C. Steppa and T. L. Holch, "HexagDLy—Processing hexagonally sampled data with CNNs in PyTorch," *SoftwareX*, vol. 9, pp. 193–198, 2019, doi: 10.1016/j.softx.2019.02.010.

[45] S. Zhang, D. Zhu, X. Yao, X. Cheng, H. He, and Y. Liu, "The Scale Effect on Spatial Interaction Patterns: An Empirical Study Using Taxi O-D data of Beijing and Shanghai," *IEEE Access*, vol. 6, no. November, pp. 51994–52003, 2018, doi: 10.1109/ACCESS.2018.2869378.

[46] X. Cheng, Z. Wang, X. Yang, L. Xu, and Y. Liu, "Multi-scale detection and interpretation of spatio-temporal anomalies of human activities represented by time-series," *Comput. Environ. Urban Syst.*, vol. 88, no. February, p. 101627, 2021, doi: 10.1016/j.compenvurbsys.2021.101627.

[47] K. F. Chu, A. Y. S. Lam, and V. O. K. Li, "Deep Multi-Scale Convolutional LSTM Network for Travel Demand and Origin-Destination Predictions," *IEEE Trans. Intell. Transp. Syst.*, vol. 21, no. 8, pp. 3219–3232, 2020, doi: 10.1109/TITS.2019.2924971.

[48] S. Openshaw, "The modifiable areal unit problem," Concepts Techn. Mod. Geogr., 1984.

[49] D. Brockmann, L. Hufnagel, and T. Geisel, ``The scaling laws of human travel,'' Nature, vol. 439, pp. 462-465, Jan. 2006.

[50] D. Zhu, N. Wang, Y. Liu, and L. Wu, "Street as a big geo-data assembly and analysis unit in urban studies: A case study using Beijing taxi data," Appl. Geogr., vol. 86, pp. 152-164, Sep. 2017.

[51] M. F. Goodchild, "Models of scale and scales of modelling," in Modelling Scale in Geographical Information Science, N. J. Tate and P. M. Atkinson, Eds. Chichester, U.K.: Wiley, 2001, pp. 3-10.

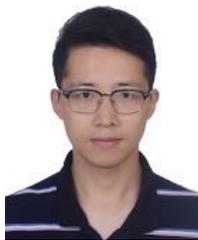

**Zhiju Chen** received the B.S. degree in traffic engineering and the master's degree in structural engineering from Zhengzhou University, in 2013 and 2018, respectively. He is currently pursuing the Ph.D. degree with the School of Transportation and Logistics, Dalian University of Technology. His research interests include machine learning and data mining in transportation.

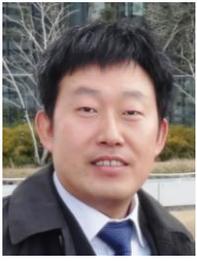
**Kai Liu** received the Ph.D. degree from Nagoya University, Japan, in 2006. He is currently a Professor with the School of Transportation and Logistics, Dalian University of Technology. His research interests include travel behaviour analysis, GIS in transportation, intelligent transport systems, and demand responsive transport system.

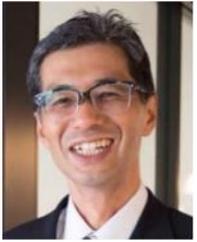
**Toshiyuki Yamamoto** received the Ph.D. degree from Kyoto University, Japan, in 2000. He is currently a Professor with the Institute of Materials and Systems for Sustainability, Nagoya University, Japan. His research interests include vehicle ownership and use, vehicle sharing systems, travel behaviour analysis, time use and activity-based analysis, intelligent transport systems, and traffic safety.

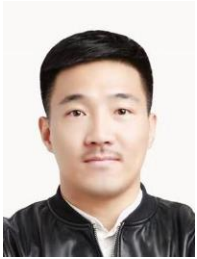
**Liheng Tuo** graduated from the University of Chinese Academy of Sciences in 2013.
He was an associate researcher in the Institute of acoustics, Chinese Academy of Sciences, in 2015-2018. He is now a senior technical expert in Didi Chuxing. The main research interest is Smart Transportation and Artificial Intelligence.